# Agentic AI Process Observability: Discovering Behavioral Variability⋆

Fabiana Fournier[1], Lior Limonad[1] and Yuval David[1]

[1]*IBM Research, Israel*

**Abstract**

AI agents that leverage Large Language Models (LLMs) are increasingly becoming core building blocks of modern software systems. A wide range of frameworks is now available to support the specification of such applications. These frameworks enable the definition of agent setups using natural language prompting, which specifies the roles, goals, and tools assigned to the various agents involved. Within such setups, agent behavior is non-deterministic for any given input, highlighting the critical need for robust debugging and observability tools. In this work, we explore the use of process and causal discovery applied to agent execution trajectories as a means of enhancing developer observability. This approach aids in monitoring and understanding the emergent variability in agent behavior. Additionally, we complement this with LLM-based static analysis techniques to distinguish between intended and unintended behavioral variability. We argue that such instrumentation is essential for giving developers greater control over evolving specifications and for identifying aspects of functionality that may require more precise and explicit definitions.

## 1. Introduction and Motivation

Artificial intelligence is advancing swiftly, transitioning from basic task automation to the development of sophisticated, autonomous systems. A key development in this progression is the emergence of Agentic AI. "This concept refers to AI systems that can perceive their environment, reason, plan, and act to achieve specific goals, much like human agents." [1]. In most contemporary frameworks realizing Agentic AI systems (e.g., CrewAI[2], LangGraph[3], AutoGen[4]), an Agentic AI is essentially a configurable wrapper around a Large Language Model (LLM) [1], typically specified with a role, goal, and behavioral constraints that enable it to operate independently or collaboratively with other agents in a shared space.

Observability is an important characteristic of an Agentic AI system, reflecting the ability to infer actionable insights about its inner workings by analyzing the inputs and outputs as

---

*PMAI25: Process Management in the AI era, Oct 25, 2025, Bologna, Italy*

⋆This project has received funding from the European Union's Horizon research and innovation programme under grant agreements no 101092639 (FAME), and 101092021 (AutoTwin).

fabiana@il.ibm.com (F. Fournier); liorli@il.ibm.com (L. Limonad); yuval.david@ibm.com (Y. David)

[1]https://bloorresearch.com/2025/02/the-rise-of-agentic-ai-the-next-step-in-artificial-intelligence/
[2]https://docs.crewai.com
[3]https://www.langchain.com/langgraph
[4]https://microsoft.github.io/autogen/stable//index.html

they are transformed and exchanged by and between the system's agents [2]. A key aspect of enabling observability is providing tools that help developers of Agentic AI applications create high-quality code in minimal time. One of the main drawbacks of this type of application is the stochastic nature of AI agents. Due to their nondeterministic behavior, sometimes the same input triggers a variety of execution trajectories yielding different results. By an agent *trajectory*, we refer to the path carried out by an agent during execution, from input to final output, including its decisions, actions, and intermediate results. Within such trajectories, there is a need to distinguish between behavior variability that is explicitly intended, hence specified, by the developer versus variability that accidentally arises during execution. Identification of the latter is important since it may undermine performance and require the developer to better crystallize the specification.

In this work, we aim to provide software engineers who develop AI agents with the means to examine points of variability arising in the specification of their agentic applications employing LLM-based static analysis. We consider agent execution trajectories as process event logs that constitute timestamped events (e.g., tool invocations) as the data source for analysis. This allows using Process Mining [3] and Causal Process Discovery [4, 5] capabilities to reveal invocation dependencies and to recognize variability that arises as split points in such views.

Following the terminology presented in [6], processes with similar inputs and outcomes can be considered variations of a single process and are referred to as *variants*. In the process model, each branching point is either a *variation point* or a *decision point*. It is a variation point if its branches correspond to different process variants; otherwise, it is a decision point. In the context of our work, we adopt this terminology to distinguish between *intended* variability, which arises from explicit decision statements in the agent specifications (i.e., decision points), and *unintended* variability, which results from the non-deterministic nature of LLM agents leading to inconsistent execution trajectories (i.e., variation points).

The work presented here represents early efforts in the emerging area of agent observability. Our main contribution lies in treating agent execution trajectories as the target of process mining. This perspective enables the use of causal and process discovery techniques to explore the behavior and collaboration of AI agents. As part of our approach, LLM-based static analysis complements the discovery process by providing additional insights into behavioral variability.

The remainder of the paper is organized as follows. Section 2 reviews related work. Section 3 introduces an example application of a calculator in CrewAI. Section 4 details our overall method for agentic process observability, which is then instantiated in the context of the example application, with results presented in Section 5. We conclude with key insights and future research directions in Section 6.

## 2. Related Work

Agentic Business Process Management (Agentic BPM) traces its roots to early work at the intersection of Multi-Agent Systems (MAS) and BPM [7]. In these early efforts, agent-centric data abstractions helped reshape complex system behavior specifications by partitioning them into smaller, encapsulated components that were easier to specify and verify. Since the introduction of the artifact-centric approach [8], this line of work has progressively laid the groundwork for process mining across multiple behavioral dimensions [9], and more recently, has initiated discussions around an Object-Centric Event Data (OCED) [10] standard. Today, with the widespread adoption of AI and the rise of LLMs, Agentic AI is experiencing a renaissance in BPM, reflected in the AI-Augmented Business Process Management Systems (ABPMS) manifesto [11], and echoed in the development of AI agent-centric BPM systems, namely Agentic BPM [12].

In this work, we employ process and causal mining in the scope of Agentic AI for the sake of process observability. The former is a relatively mature discipline with instances of an agentic flavor already explored with the goal of specification verification [7]. The application of causal mining to Agentic AI is relatively new.

Causal discovery aims to uncover causal relationships from observational data, distinguishing cause-effect directionality from mere correlation [13, 14]. Previous work on causal discovery from process data generated graphs merely based on key performance indicators [15] or decision points [16, 17, 18]. Our method for causal discovery in business processes [4, 5] infers causal graphs from the activity timestamps by adapting and extending the work in [19]. We employ causal discovery over agent execution trajectories to uncover function calls and tool invocations dependencies within and between the agents. The novelty of our work lies in leveraging process mining and causal process discovery, based on the execution times of activities, to identify variability in the non-deterministic behavior of AI agents. More specifically, it aims to reveal invocation dependencies and recognize variability that arises at split points in these views.

The types and the configuration of LLMs employed by agents can significantly influence agent behavior. Although parameters such as temperature, top-k, top-p, and repetition penalty are commonly used to reduce non-deterministic responses to identical or similar inputs, recent work already concludes that even with stricter settings, such as setting the temperature to zero, LLMs can still exhibit notable instability [20, 21]. Consequently, observability of such behavioral variabilities is crucial, not only for selecting among different LLM models to be associated with different agents, but also for guiding developers in 'tightening' all loose ends in the agent specifications, ultimately supporting a more consistent user experience.

The concept of variability has been extensively studied in Software Engineering (SE), particularly in the context of feature modeling within the paradigm of software product lines [22]. In this paradigm, variability is seen as a means of introducing flexibility into the software architecture, enabling multiple alternative instantiations of a single specification to suit different deployment needs. A similar concept was adopted in BPM, as in [23], where variability denotes customizable elements in a process model representing a family of business process variants. In our work, by contrast, we focus on the undesired form of variability that arises accidentally due to insufficiently rigorous specifications. These 'loose ends' in the design enable agents to perform unforeseen behaviors during execution.

We leverage an LLM-based static analysis approach to highlight the sources of variability in the specifications. Traditionally, static analysis is an integral component of SE and involves examining source code without executing it, to identify potential errors, code quality issues, and security vulnerabilities [24]. Given the natural language style in which AI agents are currently specified, captured by the recently coined term *vibe coding*[5], our approach aligns with recent work leveraging LLMs for static analysis in SE [25, 26, 27].

---

[5]https://x.com/karpathy/status/1886192184808149383

```python
calculation_task = Task(
        description=f"""
        Use the provided operations to calculate the result of the expression.

        For each operation in the sequence:
        1. If an operand is a variable (like E0), substitute its current value
        2. Use the appropriate tool to perform the calculation:
           - addition(a, b)
           - subtraction(a, b)
           - multiplication(a, b)
           - division(a, b)
           - evaluate_parentheses(expr)
        3. Store the result in the variable specified by "name"

        For every calculation step, show:
        - The operation being performed: "[name] = [operation]([op1], [op2])"
        - The tool being used with resolved values: "Using tool: [tool_name]([value1], [value2])"
        - The result: "Result: [value]"

        IMPORTANT:
        - You MUST use the exact tool matching the operation
        - You MUST show your work for each step
        - You MUST substitute variable values correctly
        - If you have multiple mathematical operations you should execute the calculation in the
          ↪ following order: First do Multiplication then Division then Addition and lastly
          ↪ Subtraction

        Return only the final numerical result at the end.
        """,
        expected_output="The calculated result as a number",
        agent=calculator_agent
    )
crew = Crew(
        agents=[decomposer_agent, calculator_agent],
        tasks=[decomposition_task, calculation_task],
        verbose=True,
        process=Process.hierarchical,
        manager_agent=manager,
        tools=math_tools
    )
# Define the agents
decomposer_agent = Agent(
    role="Expression Decomposer", goal="Decompose the given expression into a sequence of operations",
      ↪ backstory="""You are a mathematical expression decomposer. Your job is to take a mathematical
      expression and break it down into a sequence of simple operations that can be calculated
      step by step. You follow PEMDAS rules and assign variables to intermediate results. You never
      ↪ calculate values - you only identify the operations needed.""", llm=llama-3-3-70b-instruct,
      ↪ verbose=True, allow_delegation=False
)
calculator_agent = Agent(
    role="Calculator", goal="Calculate expressions using only the provided tools", backstory="""You
      ↪ are a calculator that can only work by using tools. For every mathematical operation, you must
      ↪ use the corresponding tool. You carefully track variables and substitute their values when
      ↪ needed.""", llm=llama-3-3-70b-instruct, verbose=True, allow_delegation=False, tools=math_tools,
      ↪ temperature=0.1
)
manager = Agent(
    role="Project Manager", goal="Efficiently manage the crew and ensure high-quality calculation
      ↪ completion, you are not allowed to call tools only to delegate work to other agents",
      ↪ backstory="You're an experienced calculation manager, skilled in overseeing complex
      ↪ calculations and guiding teams to correctly compute mathematical formulae. Your role is to
      ↪ coordinate the efforts of the crew members, ensuring that each task is completed on time and
      ↪ to the highest standard. but you do not call the tools yourself only to your agents",
      ↪ allow_delegation=True, llm=llama-3-3-70b-instruct
)
```

**Figure 1:** CrewAI Calculator App specifications

## 3. Example Application

We use a simple toy example of a calculator application in CrewAI as shown in Figure 1 to evaluate basic calculations when given mathematical formulae as input. In its setup, three agents were explicitly defined, *Decomposer*, *Calculator*, and *Manager*. The Calculator agent was assigned a calculation task, and the Decomposer agent was assigned a decomposition task relevant to expressions with parentheses. The Manager agent oversees the overall execution process and alters the delegation of responsibilities between the other agents. To fulfill the calculation task, a set of math tools is made available, including multiplication, division, addition, subtraction, and parentheses evaluation.

## 4. Approach

Overall, our approach is the process depicted in Figure 2. It facilitates an ongoing, high-level *create–insight–improve* development cycle as the code is being shaped by the developer. The process consists of the following steps:

1. Trajectory files generation – A set of #k runs is invoked with a given input as a basis for the analysis. In each run, the full execution trajectory of the agents is captured in a corresponding log file, recording every agent action, particularly tool invocations, along with its associated timestamp. For this step, we conducted 290 runs of the Calculator application using the same input. Specifically, we used the formula $1 + 2 - 3 * 4/5$, chosen to ensure that each of the basic arithmetic operations appears exactly once.

2. Event-log processing – A single consolidated process event log is compiled from the trajectory files, the 290 trajectory logs in our case. This processing step extracts the tool invocations performed by each agent, along with their corresponding timestamps. The resulting data is organized as a tabular event-log structure, where each row represents a single, timestamped invocation of a tool by an agent. From a process mining perspective, the log is examined by using the concatenation of the agent and tool columns as the activity type identifier and the run number as the trace identifier.

3. Process and causal discovery - subsequently, process and causal discovery are applied as two complementary views to form the collective execution flows. More specifically, the causal view depicts functional dependencies (among tool invocations) and variability via logical gateways [5] (i.e., split points in this view), whereas the process view captures temporal dependencies and frequencies. We employed Heuristics [28] and Causal [4, 5] Process Mining on the input event log.

4. Rule derivation – For each split point identified as a gateway within the causal model, the developer can examine its essence (i.e., whether it represents a variation or decision point). To support this, rule derivation is applied, producing a rule statement for the selected gateway that captures the control-flow structure it represents within the causal model. In our example, we chose the $XOR\_0$ gateway to demonstrate this. In a more realistic scenario, it is likely that the developer will address all other split points.

5. Static analysis - For any selected gateway, we apply LLM-based static analysis to distinguish decision points from variation points. Given the corresponding rule statement as input, the LLM is prompted to match it against the source text of the agentic application to identify its manifestation. If a matching instruction is found in the application specification, it is highlighted for the developer's attention. Details about the choice of LLM input and output prompts are elaborated below in our example application results.

6. Reliability calculation - Complementing the gateway selection, our approach also includes a statistical computation of reliability based on the frequencies in the process model to determine the number of runs required and the degree of confidence, as elaborated below.

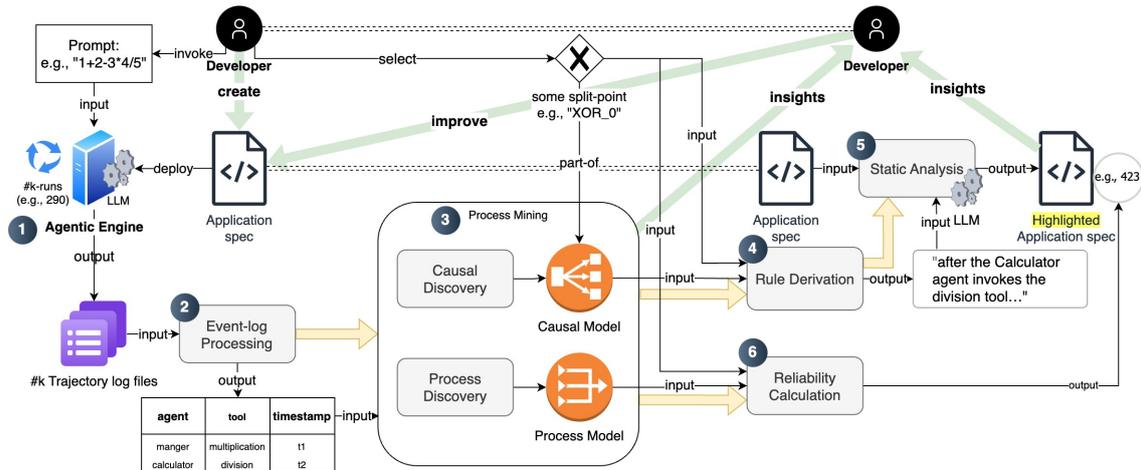

**Figure 2:** Agentic AI Process Observability Approach: Steps are numbered, yellow arrows indicate the main sequence, and green arrows depict the high-level *create–insight–improve* development cycle.

## Reliability Calculation for Gateways

The analysis of gateways also raises the question of whether the data acquired is sufficient to infer faithful conclusions about each variation point, considering the number of observation runs traversing it and the proportions of observed outbound runs.

Drawing on the normal approximation to the binomial distribution, the minimum required sample size to estimate an observed proportion $p$ of process runs following a specific branch at a gateway is given by the formula $n = \frac{Z^2 p \cdot (1-p)}{E^2}$, where $n$ is the required sample size, $Z$ is the Z-score corresponding to the desired confidence level (e.g., $Z = 1.96$ for 95% confidence), $p$ is the observed branch proportion, and $E$ is the desired margin of error (e.g., $E = 0.05$ for $\pm 5\%$). To estimate the observed proportion, a pilot sample is required. In our case, we used the initial set of 290 runs for this purpose.

This should be complemented by ensuring sufficient sampling to detect rare branches that may not yet have been observed. Specifically, to be 95% confident that a branch with true prevalence $p$ is observed at least once, the number of required runs $n$ must also satisfy $(1-p)^n < 0.05$.

Overall, for any given gateway, the number of observed runs must exceed both the minimum required to estimate observed branch proportions accurately and the threshold necessary to detect unobserved rare branches with high confidence. Complementing our last step, we also pursue an analysis of the minimal number of runs.

## 5. Example Application Results

Applying our approach to the calculator example yielded the following results. For the given input, the two execution graphs are illustrated in Figure 3.

The heuristics view (Figure 3A) helps trace outlier (e.g., less frequent) trajectories. In our example application, it shows an unusual loophole invocation of the "Evaluate_parentheses" tool by the Calculator agent despite the fact that there were no parentheses in the input. From analyzing the logs, we discovered that this behavior was triggered by the LLM arbitrarily surrounding some sub-expressions with parentheses.

The causal view (Figure 3B) depicts the invocation tool calls associated with each agent, helping to identify possible 'breaches of responsibility', having an agent invoke a tool that does not correspond to its role according to the application specification. In our case, the Project Manager agent was not explicitly granted access to any of the *math_tools*. However, in the majority of the execution trajectories, it invoked these operations directly without delegation. In addition, the causal view also captures the variability in the execution of the trajectories as illustrated by the diamond-shaped gateways. This serves our further exploration of the concrete type of each of these gateways. For static analysis, we populated the prompt, which also included the rule statement, to describe the junction structure of the $XOR\_0$ gateway as highlighted in Figure 4. This prompt was presented to an LLM (LLaMA 3-3 70B Instruct), along with the application specification shown in Figure 1, for matching purposes. We used LLaMA 3-3 70B Instruct for static Python code analysis due to its strong instruction-following capabilities and demonstrated effectiveness in code understanding and generation tasks, as evidenced in recent evaluations [29]. As illustrated, the LLM's response (Figure 5) identified a part of the task specification (also highlighted in Figure 1) as implying the rule statement for the $XOR\_0$ gateway, recognizing it as a decision point. However, as also noted, this implication fails to account for the alternation of control between the two agents—an aspect not explicitly specified in the application and one that requires further attention from the application developer. With the help of the aforementioned views and the static analysis, the developer of the application is able to determine that the Project Manager agent's definition should be augmented with a *tools=[]* entry to eliminate the discovered breach of responsibility.

### Reliability Assessment of the XOR_0 Gateway

Lastly, we examined the reliability of the $XOR\_0$ gateway with respect to the minimum number of runs required. Given the current number of observed runs through this gateway, and targeting a margin of error of 5% at a 95% confidence level (i.e., $E = 0.05$ and $Z = 1.96$), each of the two branches requires approximately 157 runs. This implies that 35 additional observations are needed beyond the current 122. Otherwise, with only 122 runs, the current margin of error remains at $\pm 5.66\%$. Adding 35 more runs (to reach 157) will reduce this to $\pm 5\%$.

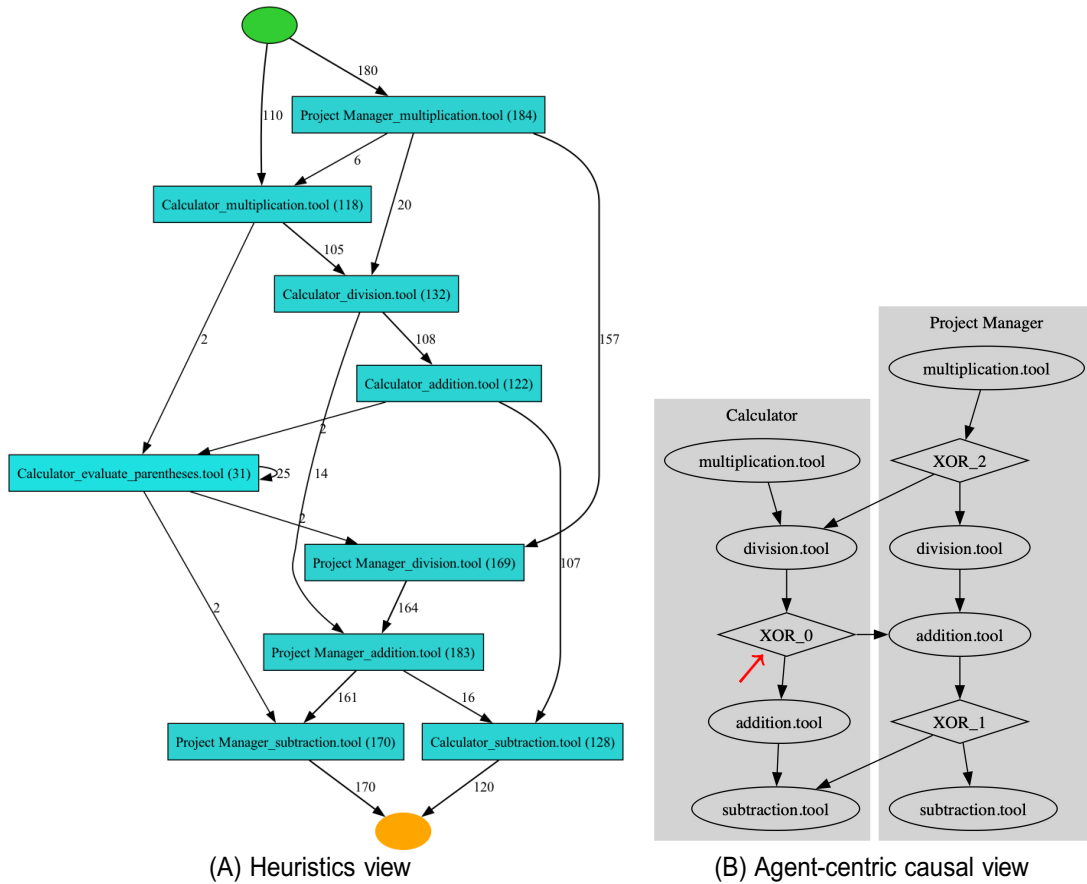

**Figure 3:** Input execution graphs for 290 runs with the input: $1 + 2 - 3 * 4/5$

```
XOR_0-matchup = f"""can you explicitly find the following rule :
"after the Calculator agent invokes the division tool, it proceeds to either the Calculator agent
↪   invoking the addition tool or to the Project Manager agent invoking the addition tool",
    in the folllowing prompt:"{prompt_for_agent}".
    dont extract the complete prompt.
    highlight an explicit statement in the prompt that is either: (1) an equivalent of the rule, (2)
↪   the statement is implied from the rule, or (3) the rule is implied from the statement
"""
```

**Figure 4:** LLM-based static analysis: searching the split logic in the specification

However, to also ensure—with 95% confidence—that a branch with a minimum true prevalence of 1% is observed at least once, the total number of runs must exceed 298. Therefore, to fully validate the gateway both in terms of proportion estimation and rare-branch detection, an additional 176 runs are required for this split, entailing 418 additional runs in total. Given the nature of our simple application example, we considered the observed margin of error of $\pm\,5.66\%$ satisfactory.

> After carefully examining the prompt, I found a statement that **implies** the rule:
> "...If you have multiple mathematical operations you should execute the calculation in the following
> ↪ order: First do Multiplication then Division then Addition and lastly Subtraction..."
> This statement implies that after performing a division operation (using the division tool), the next
> ↪ operation will be either an addition operation (which could be invoked by the Calculator agent or
> ↪ the Project Manager agent). Although it does not explicitly mention the agents, the order of
> ↪ operations suggests that after division, addition is the next step, which aligns with the rule.

**Figure 5:** Static analysis response: rule is implied but without agent responsibilities

## Modifying the specification

We adapted the specification shown in Figure 1, adding *tools=[]* to explicitly prohibit the manager agent from executing any tools. Figure 6(A) illustrates the result of this modification. As shown, while this change eliminated the manager agent's breach of responsibility, it still continued to invoke some tools.

To address this, we further revised the description of the calculation task and the manager's backstory to more strictly prohibit tool usage by the manager agent, as depicted in Figure 7. This additional refinement successfully enforced the intended behavior, restricting tool invocation solely to the calculator agent as shown in Figure 6(B).

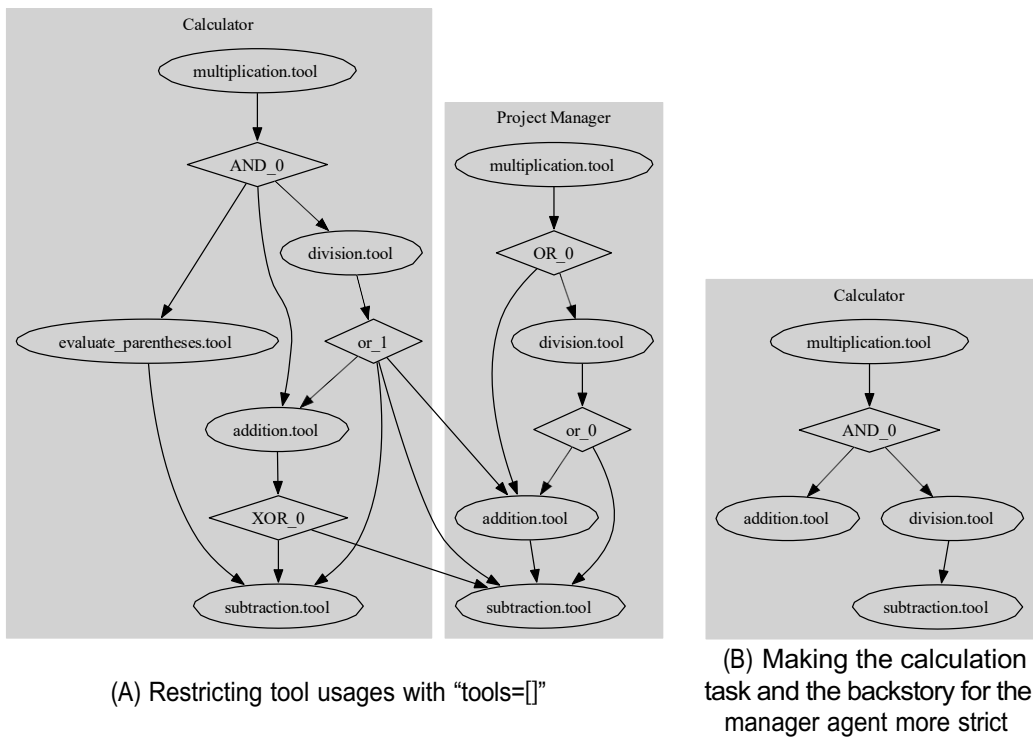

(A) Restricting tool usages with "tools=[]"

(B) Making the calculation task and the backstory for the manager agent more strict

**Figure 6:** Causal views after specification modifications

```
calculation_task = Task(
        description=f"""
            ONLY the Calculator should use tools for this task.
            The manager should delegate this entire task to the calculator_agent.
            ...
            """,
        ...
    )
manager = Agent(
    ...
    backstory="You are a project manager who coordinates work but NEVER performs calculations yourself.
    ↪  Your only job is to delegate tasks to the appropriate agents. You do not have access to any
    ↪  tools and must not attempt to use them. All actual work must be done by your team members.",
    ↪  tools=[],
    ...
)
```

**Figure 7:** Revised task and manager's backstory

## 6. Conclusions and Outlook

In this work, we propose an approach for agent observability that leverages two complementary techniques of process and causal discovery to identify points of variability in agent trajectories. We then apply LLM-based static analysis to determine the nature of these variation points. Our contribution is further complemented by a reliability measurement for split points.

We illustrated the approach using an example of a calculator application, demonstrating the possible valuable insights that such instrumentation can provide to support developers engaged in Agents DevOps. Our preliminary results show the potential of applying our framework in the context of observability. We acknowledge that further empirical validation on real-world applications and with other agentic frameworks is needed to establish the robustness of our approach. Furthermore, future work should investigate how multiple input utterances can be populated to enable joint observation and robust testing coverage.

Our control-flow-based realization for rule derivation is currently agnostic to the potential data richness underlying such decision points, and future research could extend this with data-aware analysis of these decisions.

Whether through single-agent input analysis or the cumulative investigation of multi-agent trajectories recorded over time in a running Agentic AI system, the domain of agent process observability presents a fresh playground for exploration using process and causal discovery tools developed over the past decades. As seen in the evolution of other application domains, this area may first emerge with observability tools and gradually progress toward realizing the vision of self-debugging and adaptive agents—agents that monitor their own execution, explain their actions, debug and enhance one another's behavior, and learn to evolve over time to become more reliable and autonomous.